\documentclass{article}

    \usepackage[final,nonatbib]{neurips_2023}

\usepackage[utf8]{inputenc} 
\usepackage[T1]{fontenc}    
\usepackage{hyperref}       
\usepackage{url}            
\usepackage{booktabs}       
\usepackage{nicefrac}       
\usepackage{microtype}      
\usepackage{xcolor}         
\usepackage{amsfonts,amsmath,mathtools,amsthm,bbm}
\usepackage{pdfpages}

\newtheorem{definition}{Definition}
\newtheorem{theorem}{Theorem}
\newtheorem*{theorem*}{Theorem}

\title{\emph{math-PVS}: A Large Language Model Framework to Map Scientific Publications to PVS Theories}

\author{%
  Hassen Saidi \\
  Computer Science Laboratory\\
  SRI International,\\
  Menlo Park, CA 94025 \\
  \texttt{hassen.saidi@sri.com} \\
   \And
   Susmit Jha \\
   Computer Science Laboratory \\
   SRI International, \\
   Menlo Park, CA 94025\\
   \texttt{susmit.jha@sri.com} \\
   \AND
   Tuhin Sahai \\
  Applied Sciences Laboratory \\
  SRI International \\
  Menlo Park, CA 94025.\\
   \texttt{tuhin.sahai@sri.com} \\
}

\begin{document}

\maketitle

\begin{abstract}
  As artificial intelligence (AI) gains greater adoption in a wide variety of applications, it has immense potential to contribute to mathematical discovery, by guiding conjecture generation, constructing counterexamples, assisting in formalizing mathematics, and discovering connections between different mathematical areas, to name a few.   
  While prior work has leveraged computers for exhaustive mathematical proof search, recent efforts based on large language models (LLMs) aspire to position computing platforms as co-contributors in the mathematical research process. Despite their current limitations in logic and mathematical tasks, there is growing interest in melding theorem proving systems with foundation models. This work investigates the applicability of LLMs in formalizing advanced mathematical concepts and proposes a framework that can critically review and check mathematical reasoning in research papers. Given the noted reasoning shortcomings of LLMs, our approach synergizes the capabilities of proof assistants, specifically PVS, with LLMs, enabling a bridge between textual descriptions in academic papers and formal specifications in PVS. By harnessing the PVS environment, coupled with data ingestion and conversion mechanisms, we envision an automated process, called \emph{math-PVS}, to extract and formalize mathematical theorems from research papers, offering an innovative tool for academic review and discovery.
\end{abstract}

\section{Introduction}
The growth (both in model size and size of training datasets) of transformer-based foundation models is quickly revolutionizing a large number of applications such as natural language processing, image generation and modification, and multi-modal inference. 
Recently, the National Academy of Sciences, Engineering, and Medicine released a workshop report~\cite{NASEM2023} that emphasizes the importance of constructing and deploying AI agents that aid in mathematical endeavors. The advances in machine learning (ML) and AI methods has already aided in generating theorems for outstanding conjectures and discovering new connections between disparate concepts~\cite{davies2021advancing}. 

Note that using computers to assist in generating proofs does have historical precedence. Examples include the four color theorem~\cite{appel1989every} and the proof of the existence of a chaotic attractor in the Lorenz system using interval arithmetic~\cite{tucker1999lorenz}. These efforts used computers to exhaustively search a large (combinatorial) search space and demonstrate the robustness of chaotic attractors respectively. Recent efforts, however, aim to elevate the use of computing platforms to that of an \emph{assistant} or a \emph{peer-participant} in the creative process of mathematical research. Given that large language models (LLMs) do not directly perform well on problems in logic and mathematics~\cite{lightman2023lets}, there are multiple efforts to integrate theorem proving systems with foundation models, see ~\cite{NASEM2023,first2023baldur,yang2023leandojo} and references therein.  

The majority of these efforts are focused on formalization of mathematics from the ground up, wherein the LLM proof assistants are customized to prove theorems in particular domains. A number of these efforts use datasets such as MATH, GSM8K, and datasets in~\cite{cobbe2021gsm8k,hendrycksmath2021,zheng2022minif2f} that are limited to high school grade mathematics and mathematics Olympiad questions. Our goal is to explore the use of LLMs on formalizing mathematics used at the frontier of research, and to examine the possibility of developing an LLM-based research assistant that can review mathematical reasoning in a research paper. It is widely accepted that LLMs lack even basic reasoning capability and fail at even spatial reasoning~\cite{jha2022responsible}. Consequently, we posit that any serious AI mathematics research assistant capable of reviewing research papers and assisting in proving theorems needs to be a combination of proof assistants such as prototype verification system (PVS)~\cite{PVS} and LLMs that can interface between the textual description in research papers and the formal specification in PVS. 


PVS~\cite{PVS} is a mechanized environment for formal specification and verification and consists of a specification language based on classical, typed higher-order logic, a large number of predefined theories, a type checker, and an interactive theorem prover. PVS type system supports sub-typing and dependent types. Moreover, the PVS prelude offers foundational theories. NasaLib \cite{PVSNasaLib} is a large collection of PVS libraries maintained by NASA. Currently, NasaLib consists of 63 top-level libraries, containing about 38K proven formulas in total. They encode basic mathematical theories such as topology, graph theory, group theory, and probability theory to name a few, as well as useful theories for reasoning about critical systems. 

\begin{figure}
    \centering
    \includegraphics[width=0.75\textwidth]{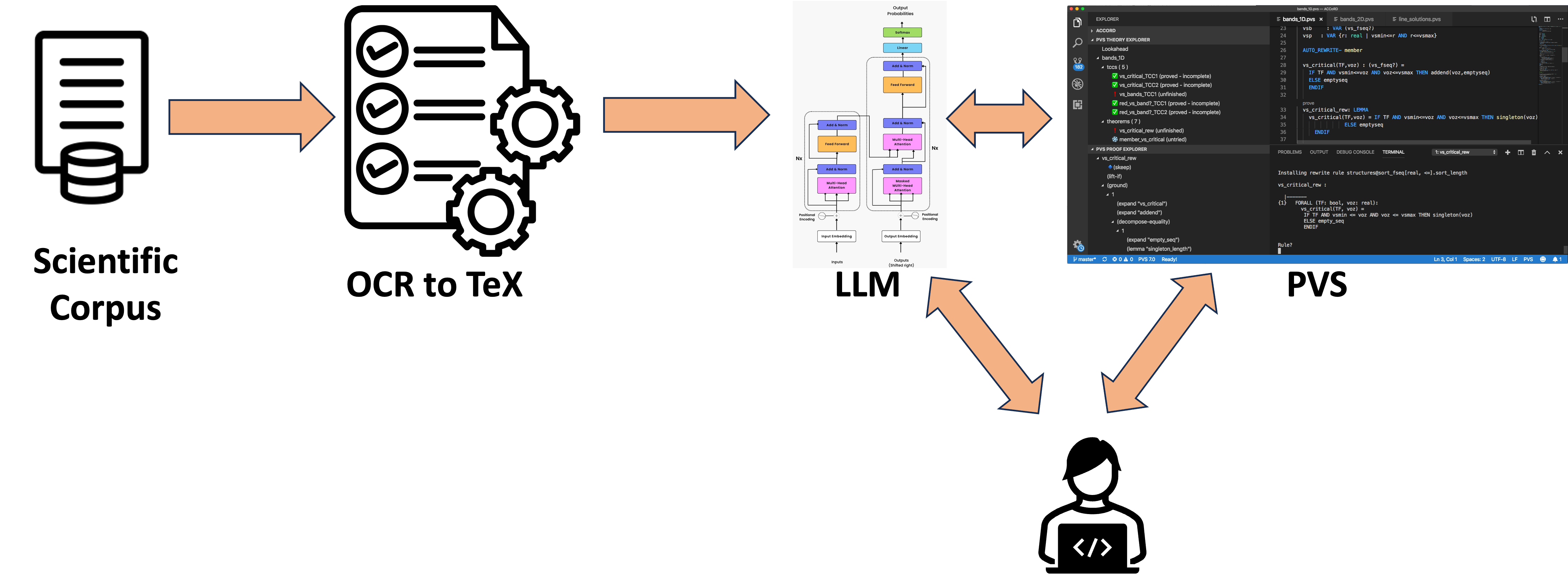}
    \caption{Overall workflow called \emph{math-PVS} for data generation from scientific publications, LLM-based PVS code generation, and interactive theorem proving.}
    \label{fig:workflow}
\end{figure}

Our framework (see Fig.~\ref{fig:workflow}) is able ingest the vast amounts of scientific literature in the form of PDF files. The ingested data is directly converted to PVS code by a LLM as follows: (a) we generate the mathematical data by processing a large corpus of scientific mathematical publications using Facebook's Nougat~\cite{blecher2023nougat} framework to convert the information from PDF to \LaTeX format, (b) the \LaTeX code is converted to PVS theories using LLMs, and (c) the user interacts with the LLM and PVS systems and iterates with them to refine the theories and proofs with the aim of discovering new proofs, define conjectures, check existing proofs, and uncover hidden connections between disparate works. This integration of the PVS with LLMs opens up the possibility of automatically extracting PVS formalization of mathematical theorems and proofs in existing research papers into PVS which can be useful paper reviews and the aforementioned tasks.  

\section{A Prompting Framework for Generating PVS Theories}
We demonstrate our framework by reproducing the core arguments for a proposition and its associated proof from an existing paper~\cite{delvenne2006decidability} that relates the evolution of Turing machines to symbolic dynamics. Note that symbolic dynamics are a standard tool in study of temporally evolving nonlinear systems. Additionally, the authors develop a generalized definition of computational universality that extends beyond Turing machines, counter machines, and tag systems. This work generalizes the notion of computational universality to cellular automata and subshifts. The authors define an effective symbolic space (ESS) and exploit it as described below.
\begin{definition}
An effective symbolic space is a pair \((X,P)\), where \(X\) is a symbolic space and \(P:\mathbb{N}\to 2^{X}\) is an injective function whose range is the set of all clopen sets of \(X\), such that the intersection and complementation of clopen sets are computable operations. This means that there exist computable functions \(f:\mathbb{N}\to\mathbb{N}\) and \(g:\mathbb{N}\times\mathbb{N}\to\mathbb{N}\) such that \(X\setminus P_{n}=P_{f(n)}\) and \(P_{n}\cap P_{m}=P_{g(n,m)}\).
\end{definition}
They show that cellular automata and Turing machines (ignoring the blank symbol) are examples of effective symbolic systems and then go on to prove universality theorems using this construct. 

We ingest the above paper (in PDF format) using the Nougat software~\cite{blecher2023nougat} and feed the resulting \LaTeX code to OpenAI's GPT-4 for automatic generation of PVS theories. In particular, we used the definition of effective symbolic spaces above along with the following definition to prove a simplified version of the first proposition in~\cite{delvenne2006decidability} (presented below for completeness). 

\begin{definition}
Let \((X,P)\) and \((Y,Q)\) be two effective symbolic spaces. An effective continuous map is a continuous map \(h:X\to Y\) such that \(h^{-1}(Q_{n})=P_{k(n)}\), for some computable map \(k:\mathbb{N}\to\mathbb{N}\). If \(h\) is bijective then it is an effective homeomorphism, and \((X,P)\) is said to be effectively homeomorphic to \((Y,Q)\).
\end{definition}

\begin{theorem}
Every effective symbolic space is effectively homeomorphic to an effective subset of the Cantor space. Every perfect effective symbolic space is effectively homeomorphic to the Cantor space.
\end{theorem}
We note that we use definitions of Cantor spaces~\cite{rudin1953principles}, homeomorphisms~\cite{guckenheimer2013nonlinear}, and other mathematical terminology from publicly available lectures notes and Wikipedia/Scholarpedia articles. In the case that these documents are available in PDF format, we again use the Nougat software to convert key portions to \LaTeX format that is then converted to PVS theories using GPT-4.

 We note that, as shown in Fig.~\ref{fig:workflow}, both the transformer-based PVS theory generation and its use in PVS presently requires human intervention. The user was also involved in simplifying the theories for PVS parsing, typechecking, and applying the {\tt grind} tactic,  a catch-all strategy that is frequently used to automatically complete a proof in PVS. Moreover, we note that these interactions were very similar to those that one may have with collaborators or advanced student researchers. 

\section{Using LLM for Abstraction}

 The human input required for our \emph{math-PVS} framework as illustrated in Figure \ref{fig:workflow} is to use LLMs to simplify and construct abstractions  of the original input that capture the underlying core concepts. Similar to the formalization of a complex system or algorithm in a theorem prover, the abstraction step is required here to map mathematical concepts described in the paper of interest, to concepts grounded in basic mathematical constructs such as set theory. While the formalization of the original input in PVS can be automated using LLMs, the resulting PVS theorems might require extensive proof details and background mathematical knowledge that are typically omitted in most mathematical papers. By using LLMs to automatically generate abstractions of the input, the resulting concepts can be mapped directly to PVS theories and subsequently proved by theorem proving techniques without the need for ingestion and processing of copious amounts of background mathematical knowledge and theories. As the corpus of mathematical knowledge grows in math-PVS, we expect this need for abstraction to reduce.

\begin{figure}[ht]
    \centering
    \includegraphics[width=0.75\textwidth]{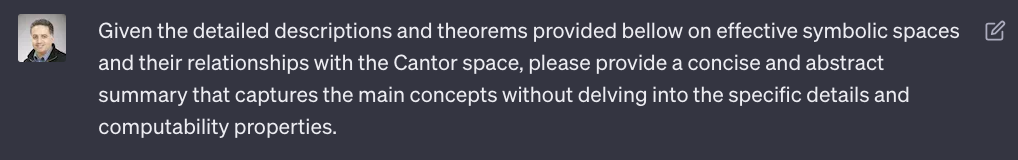}
    \caption{Prompting ChatGPT4 to provide abstraction of the original mathematical definitions and theorem}
    \label{fig:abstraction_prompt}
\end{figure}

Figure \ref{fig:abstraction_prompt} illustrates the prompt used to abstract the mathematical definitions extracted from the paper~\cite{delvenne2006decidability}. The prompt explicitly requests the LLM to generate an abstraction that captures the main concepts without the inclusion of specific details, thereby enabling the foundation model to generate a self-contained and consistent PVS code.

\section{Evaluation}

We now describe how our approach can be applied to review the paper~\cite{delvenne2006decidability} described in the previous section.
The first step in our process is to use Meta's Nougat \cite{blecher2023nougat} to generate \TeX from the original PDF file. 
Nougat is a Visual Transformer model that performs an Optical Character Recognition (OCR) task for processing scientific documents into a markup language such as \LaTeX. This makes the mathematical knowledge accessible and enables us to use LLMs for autoformalization into PVS.

\begin{figure}[ht]
    \centering
    \includegraphics[width=0.75\textwidth]{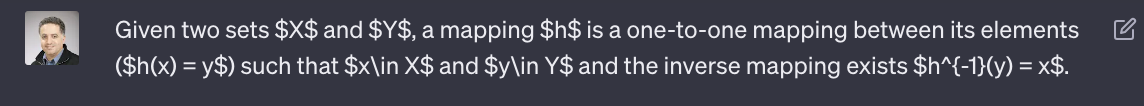}
    \includegraphics[width=0.65\textwidth]{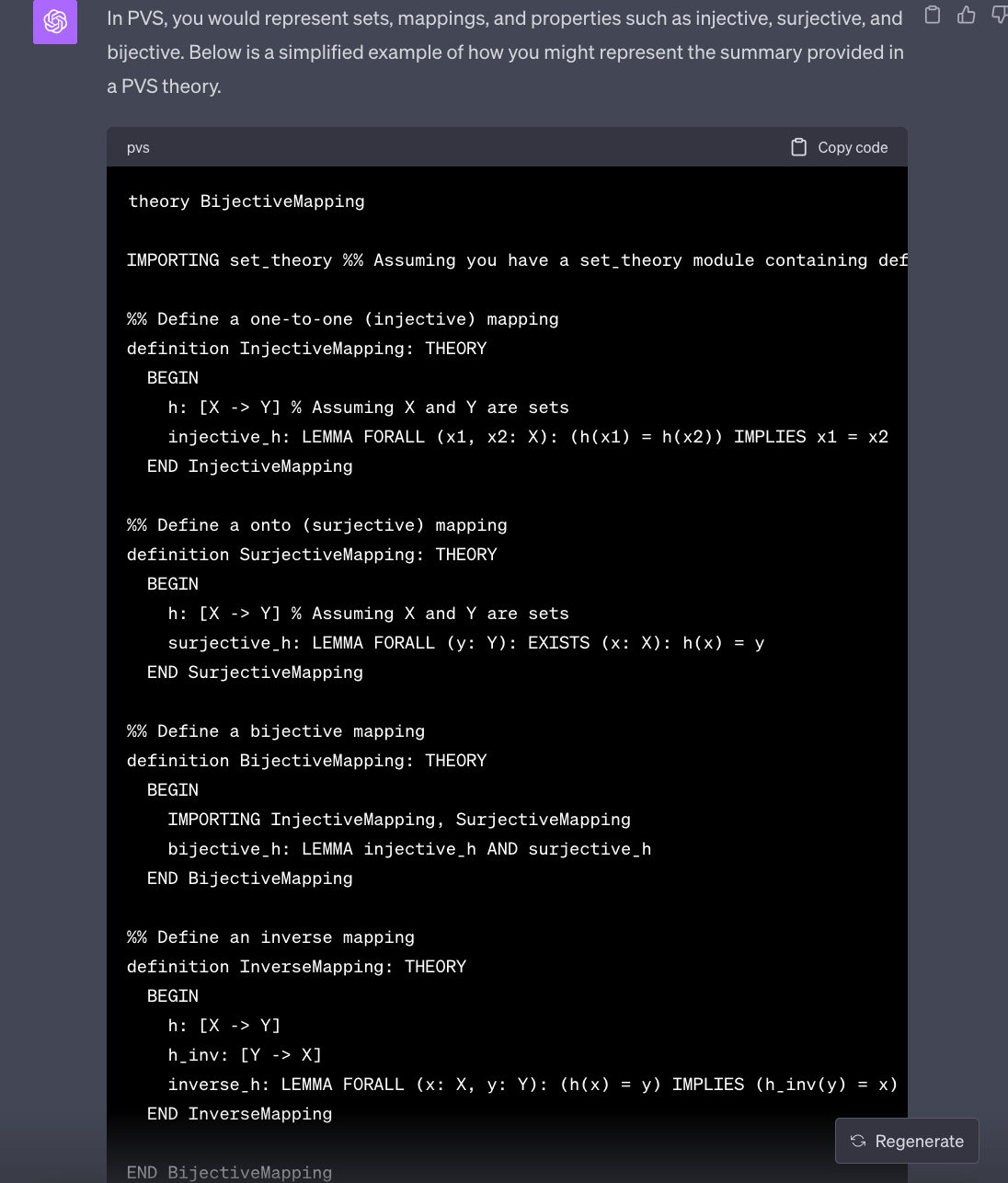}
    \caption{Prompting ChatGPT4 with a bijective mapping definition, and the resulting PVS theory}
    \label{fig:definition_prompt}
\end{figure}

The next step is to extract and summarize the definitions and theorems  from the recovered Latex. The core theory  described in the paper under study~\cite{delvenne2006decidability}  can be summarized as follows:
\begin{definition}
Given two sets $X$ and $Y$, a mapping $h$ is a one-to-one mapping between its elements ($h(x) = y$) such that $x\in X$ and $y\in Y$ and the inverse mapping exists $h^{-1}(y) = x$.
\end{definition}
\begin{definition}
A Cantor space is the set of infinite numbers on the interval $\left[0,1\right]$ that can be represented as a sequence $a_0,a_1,\hdots$ such that $a_i\in\{0,1\}$    
\end{definition}
\begin{definition}
A symbolic space is a set such that there exists a one-to-one map to some subset of integers greater than $0$. 
\end{definition}
\begin{definition}
 An effective symbolic space is a pair $(X,P)$ where $X$ is a symbolic space and $P$ is a one-to-one mapping $P:X\rightarrow 2^{X}$. Here $2^{X}$ is the power set of all possible combinations of subsets of $X$. 
\end{definition}
\begin{theorem*}
 Every effective symbolic space has a homeomorphic mapping to some subset of a cantor space.   
\end{theorem*}

We note that the above definitions are not entirely complete, however, they capture the central concepts associated with the proof that enable one to create the mappings between the spaces. This is consistent with a mathematical collaborator that is participating in the theorem proving activity. 
We proceed by simply pasting definitions and theorems from the summary as prompts without giving any context. Figure \ref{fig:definition_prompt} show how a definition of a bijective mapping is used as a prompt, and the output of ChatGPT4 when prompted to generate a PVS theory implementing the definition. For further details we point the reader to the appendix.

\noindent \textbf{Limitations.} The generated PVS theory is very close to a valid PVS theory that is syntactically correct (can be parsed) and semantically correct (can be typechecked). But the autoformalization by LLMs is not perfect and without errors. For instance, {\tt theory Mappings} should be 
{\tt Mappings: THEORY}. The importing statement should be after 
{\tt begin}. Also, {\tt set\_theory} in PVS is simply {\tt sets} defined in the prelude. These syntactic errors are easily fixed by providing ChatGPT4 with a few examples of PVS syntax. In future work, we plan to address these issues by fine tuning ChatGPT4 using the PVS prelude and the PVS NASA library \cite{PVSNasaLib} which have large quantities of PVS code. We will also explore combining this work with CoProver \cite{coprover} to automatically generate lemmas from the extracted theories, and automate proofs generation.

\section{Conclusion and Future Work}

In this work, we explored the transformative potential of integrating  LLMs and proof assistants such as PVS. By automating the extraction and formalization of mathematical content from research papers, we provide a promising approach for academic review and knowledge discovery. This fusion of AI and formal mathematical methods paves the way for more robust, collaborative research practices for scientific discovery. We envision the following uses for our LLM-based \emph{math-PVS} theorem proving framework.
\begin{enumerate}
    \item Automated reviewing of papers by journals and reviewers for theorems and their proofs.
    \item Discovery of connections between unrelated subareas of mathematics enabling the extensions of existing theories to new settings.
    \item Construction of an assistant that removes the mechanical portions of theorem proving, thereby enabling the mathematician researcher to focus on the creative aspects of the work and the ``bigger picture''. 
\end{enumerate}

In future work, we believe that several steps in the process will be automated, reducing the workload on the scientific user of the math-PVS framework. Using the extensive PVS NASA library, we plan to fine-tune a foundation model using existing PVS theories to reduce the number of errors in PVS theory code generation. 

Moreover, we plan to process a large corpus of mathematical literature using Nougat and generate extensive math-PVS theories that will reduce the workload of the user.

\bibliographystyle{unsrt}
\bibliography{AI_discovery}

\appendix
\section{Appendix: ChatGPT4 Prompts, PVS theory, and PVS Proof}

This appendix contains specific details of the example outlined in the main body of the paper. In particular, we provide the following information:
\begin{enumerate}
    \item The ChatGPT4 prompts and responses for the example reported in the main body of the paper.
    \item The PVS theory generated by ChatGPT4 for the example in the main body of the paper.
    \item edited and final PVS theory for the example in the main body of the paper.
    \item The proof of the main theorem in PVS after it has been abstracted to the core concepts.
\end{enumerate}

The first set of interactions with ChatGPT4 are used to input the original definitions and proposition from the paper, we then use ChatGPT4 to summarize the information, abstract it, and generate a graph that maps all the concepts introduced in the definitions along with their dependencies.

The second set of interactions  with ChatGPT4 is used to generate PVS theories from the definitions and summaries and prove the primary abstracted version of the proposition of interest.


\includegraphics[width=1.0\textwidth]{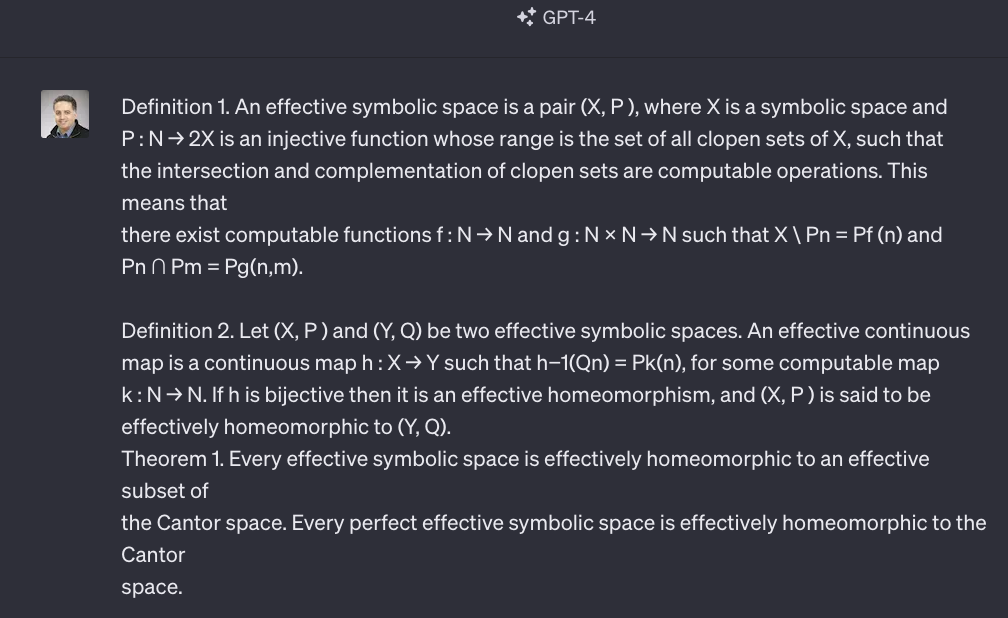}

\includegraphics[width=1.0\textwidth]{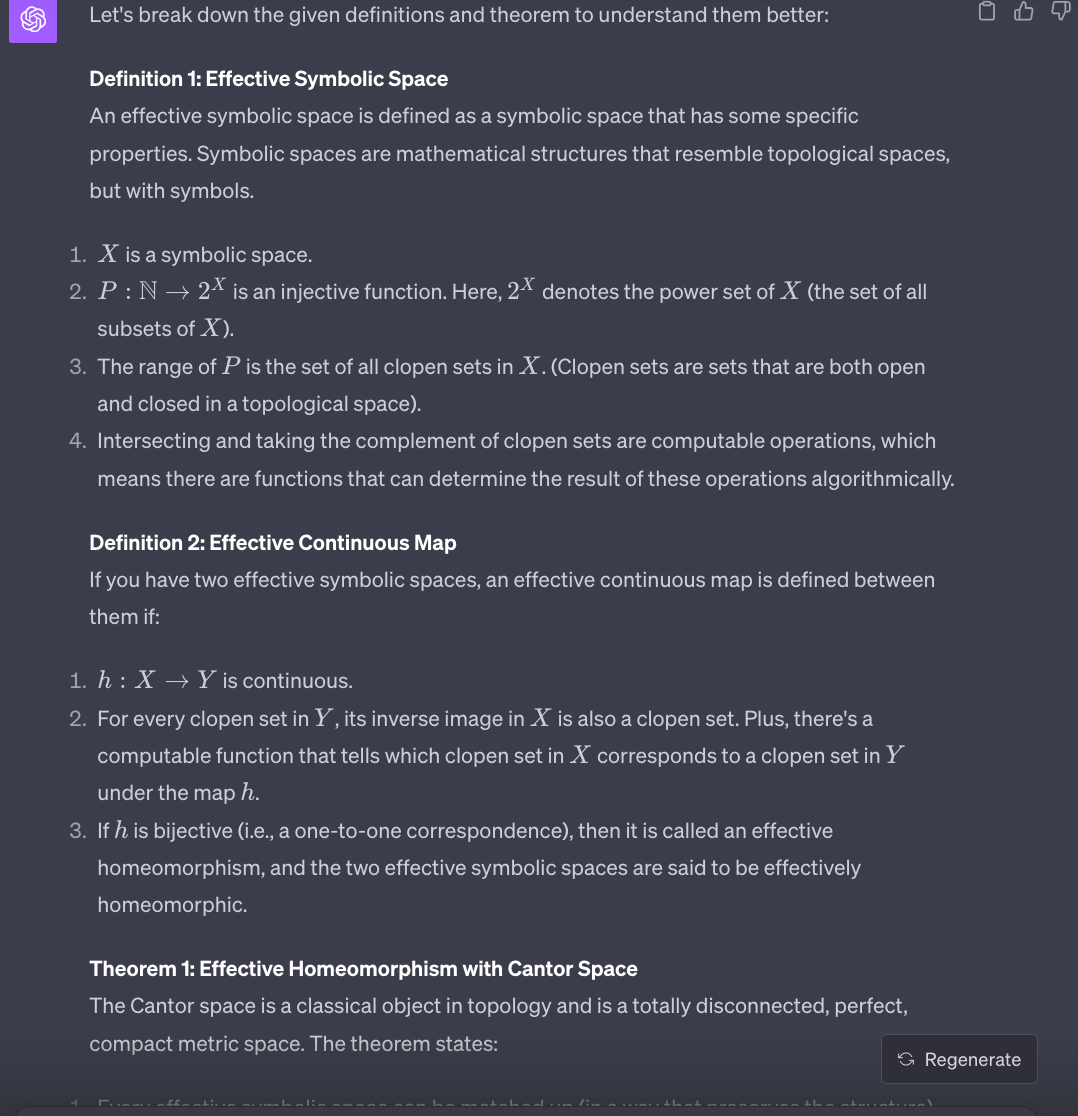}

\includegraphics[width=1.0\textwidth]{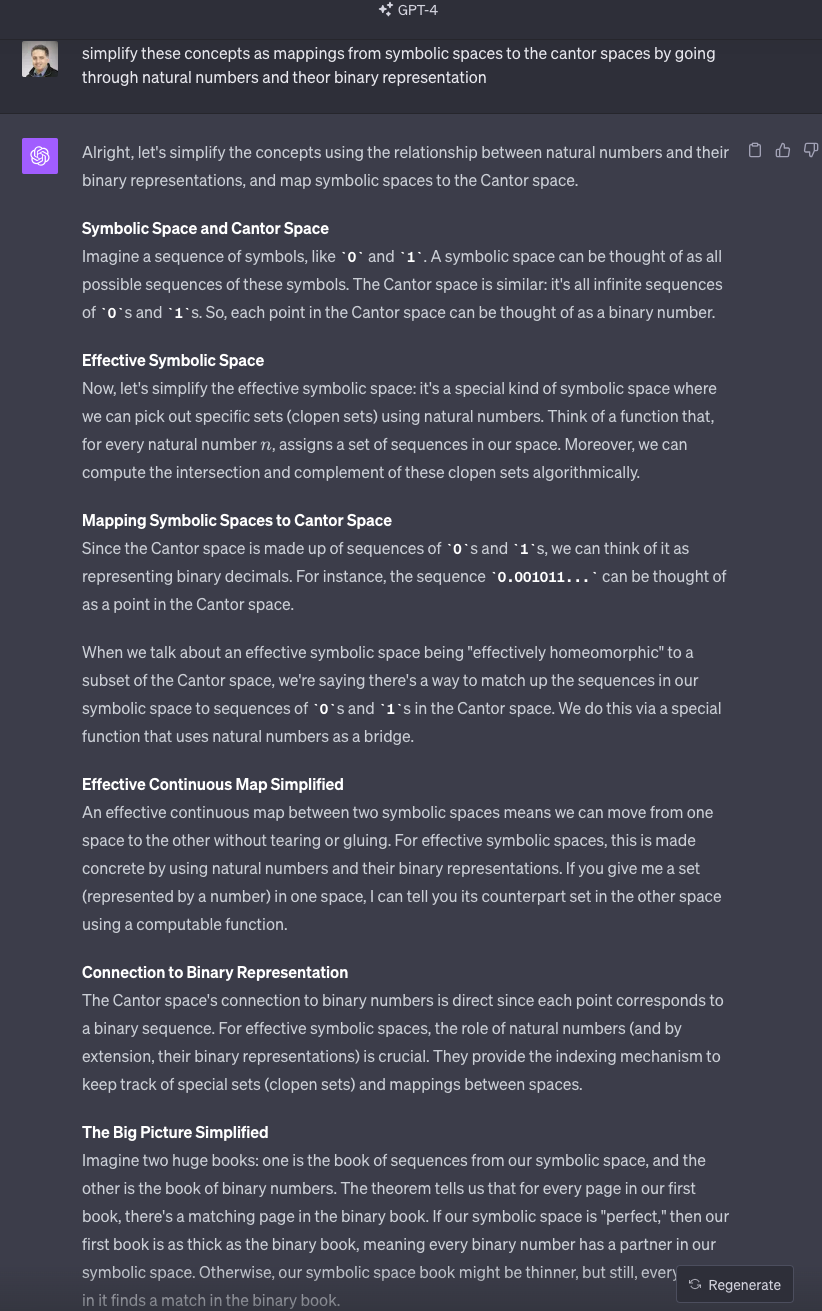}

\includegraphics[width=1.0\textwidth]{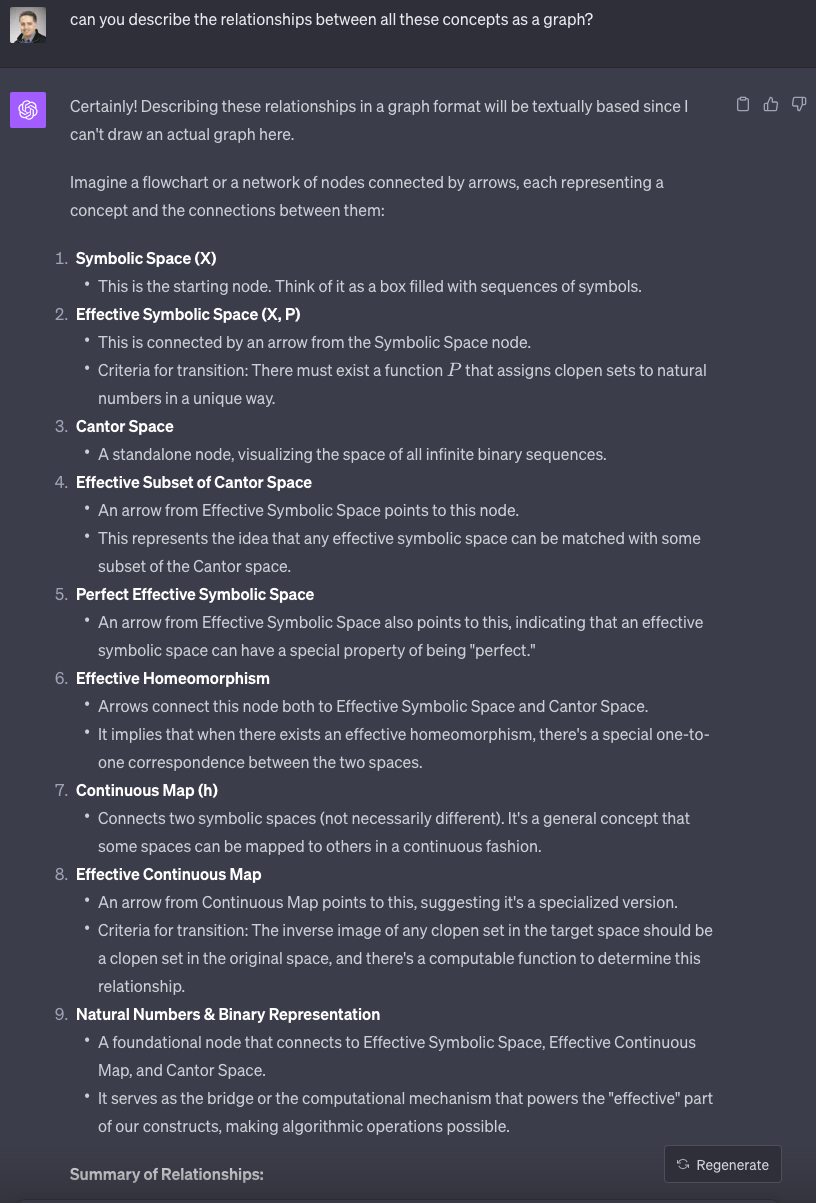}

\includegraphics[width=1.0\textwidth]{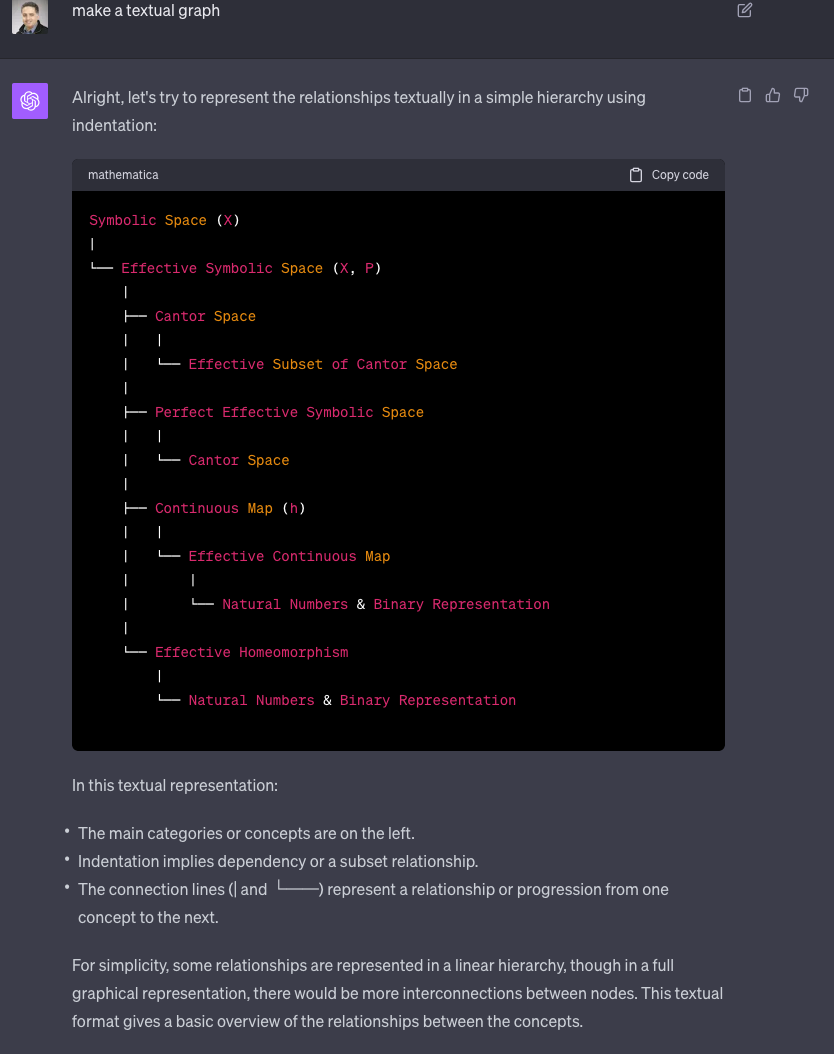}

\includegraphics[width=1.0\textwidth]{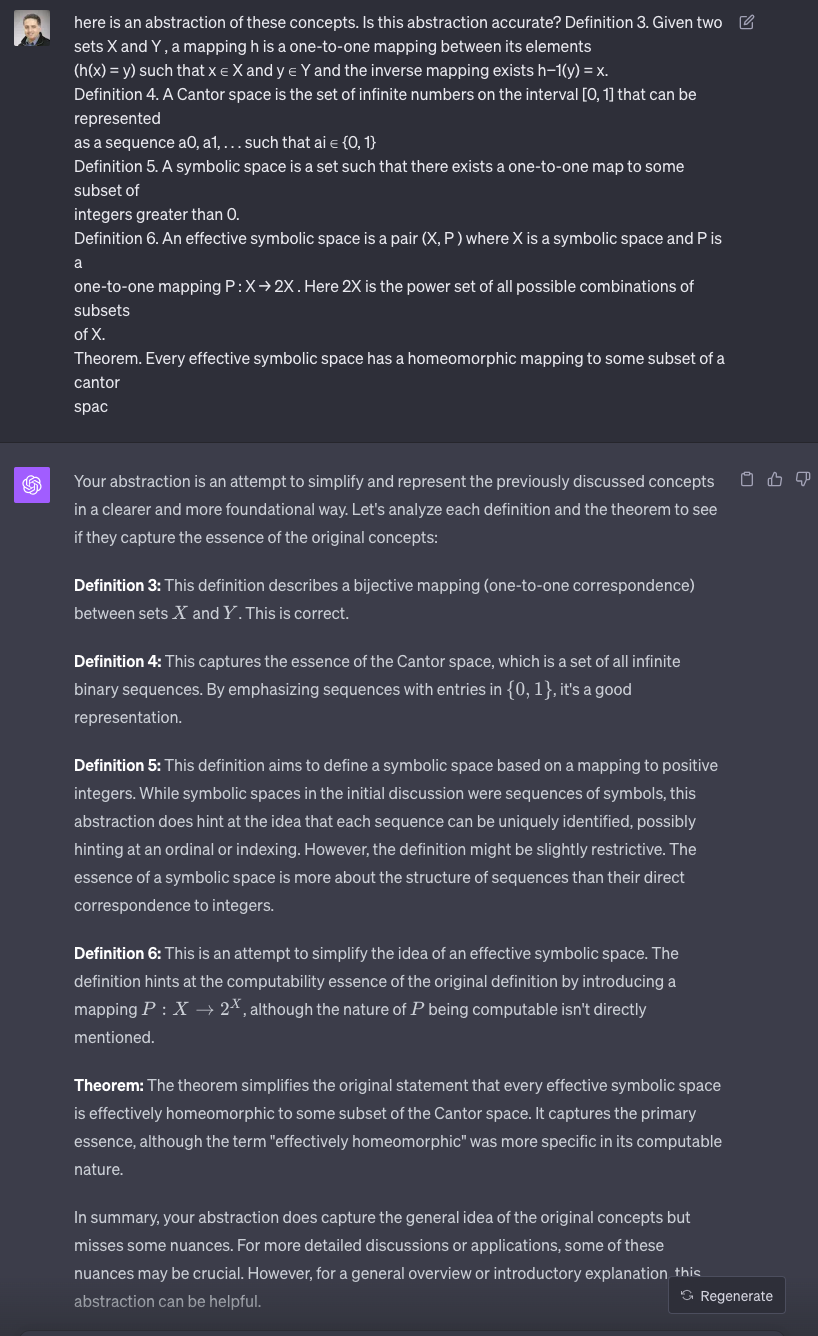}

\includegraphics[width=1.0\textwidth]{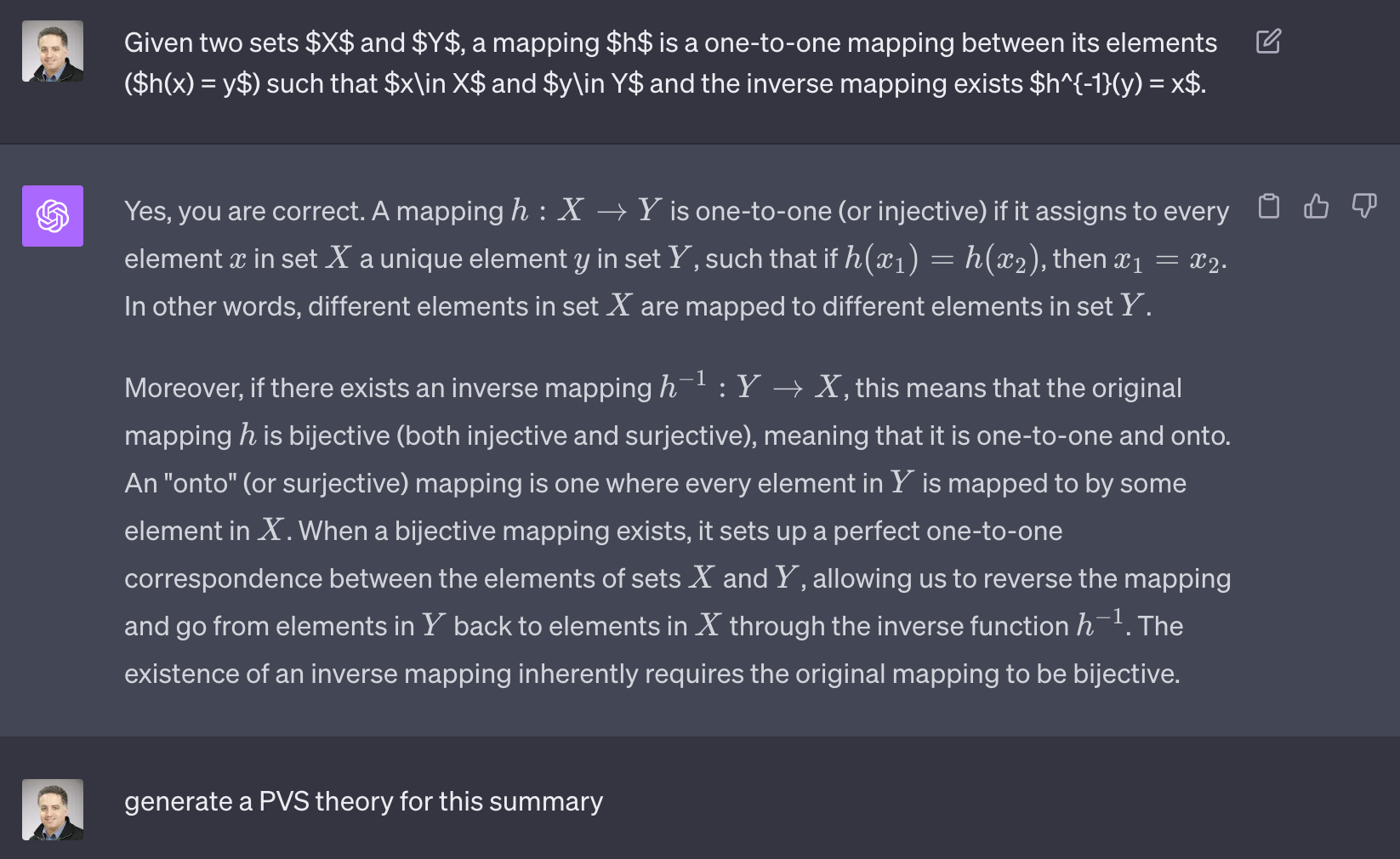}

From this point, we ask ChatGPT4 to generate PVS theories formalizing the various definitions and theorems.
The resulting PVS theories are manually edited to produce a single theory merging the various generated theories. In particular, (1) fix syntax errors in the generated PVS code, and (2) change the theorem from proving a surjective mapping to a bijective mapping, and (3) added two lemmas that allow us to prove the main result.

\includegraphics[width=1.0\textwidth]{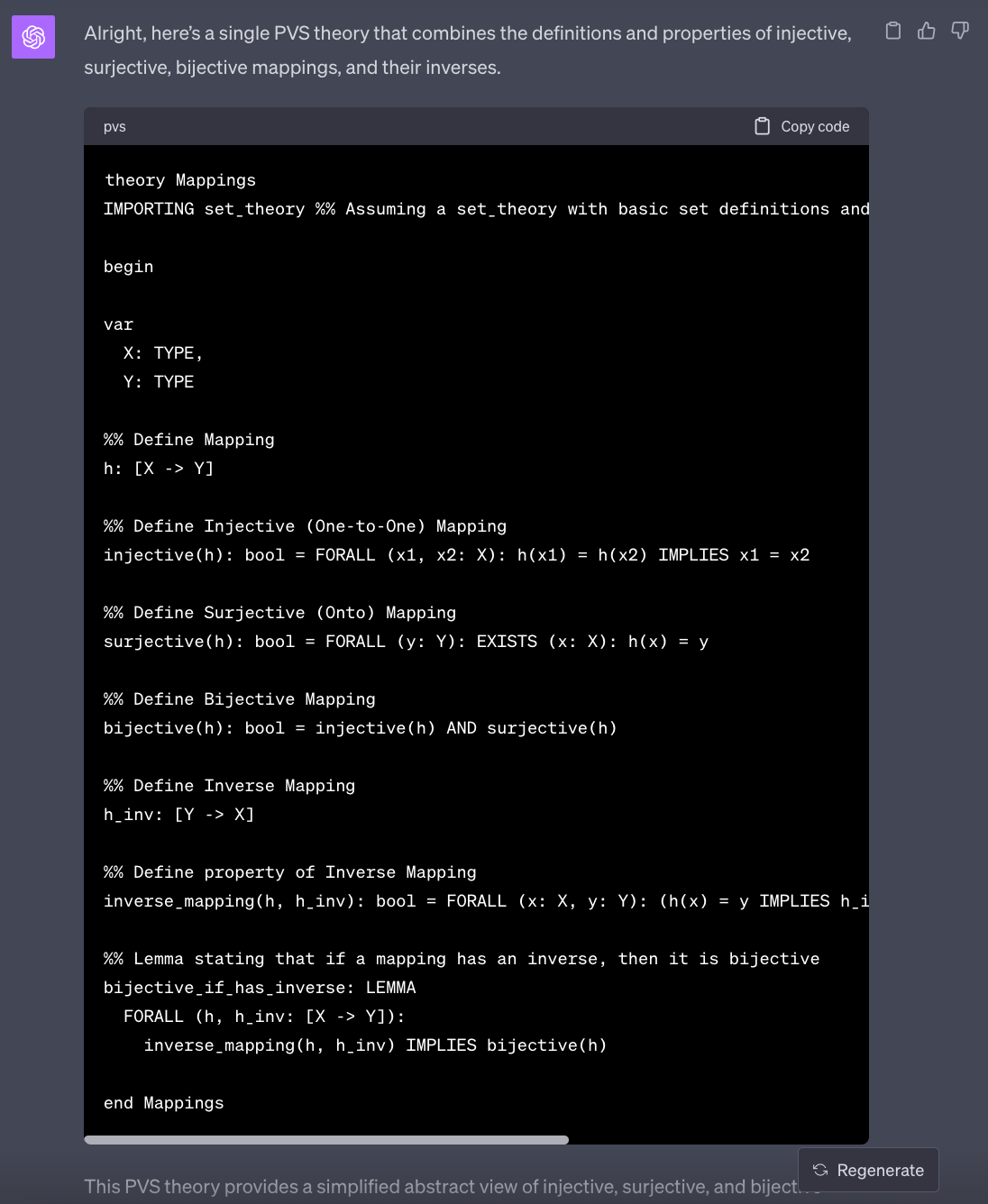}

\includegraphics[width=1.0\textwidth]{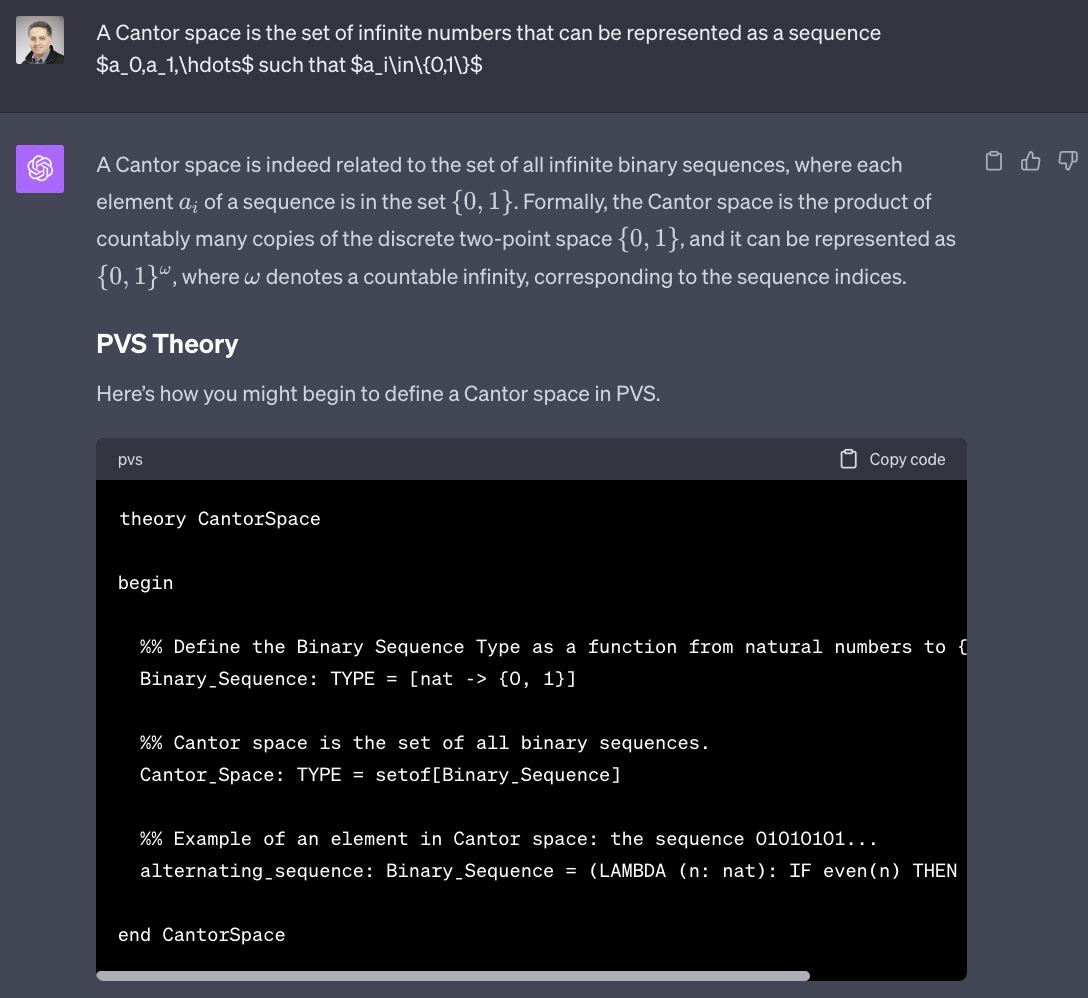}

\includegraphics[width=1.0\textwidth]{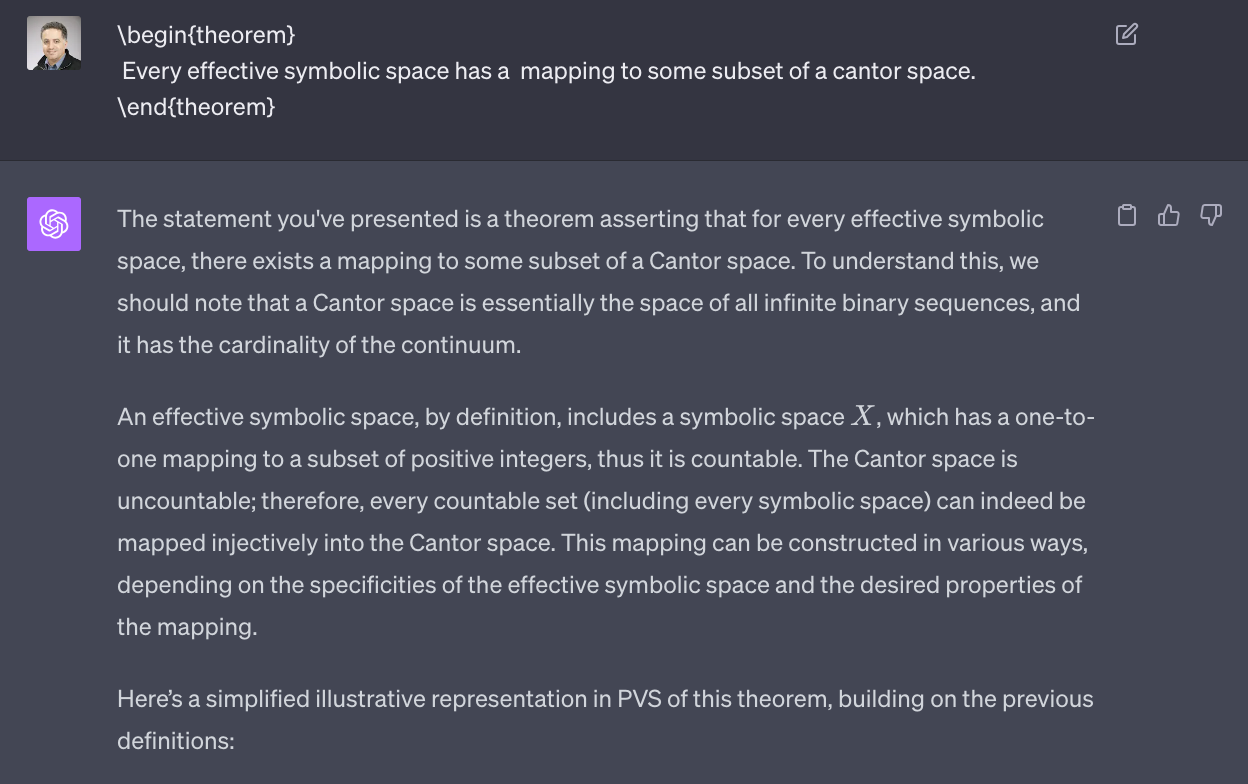}

\includegraphics[width=1.0\textwidth]{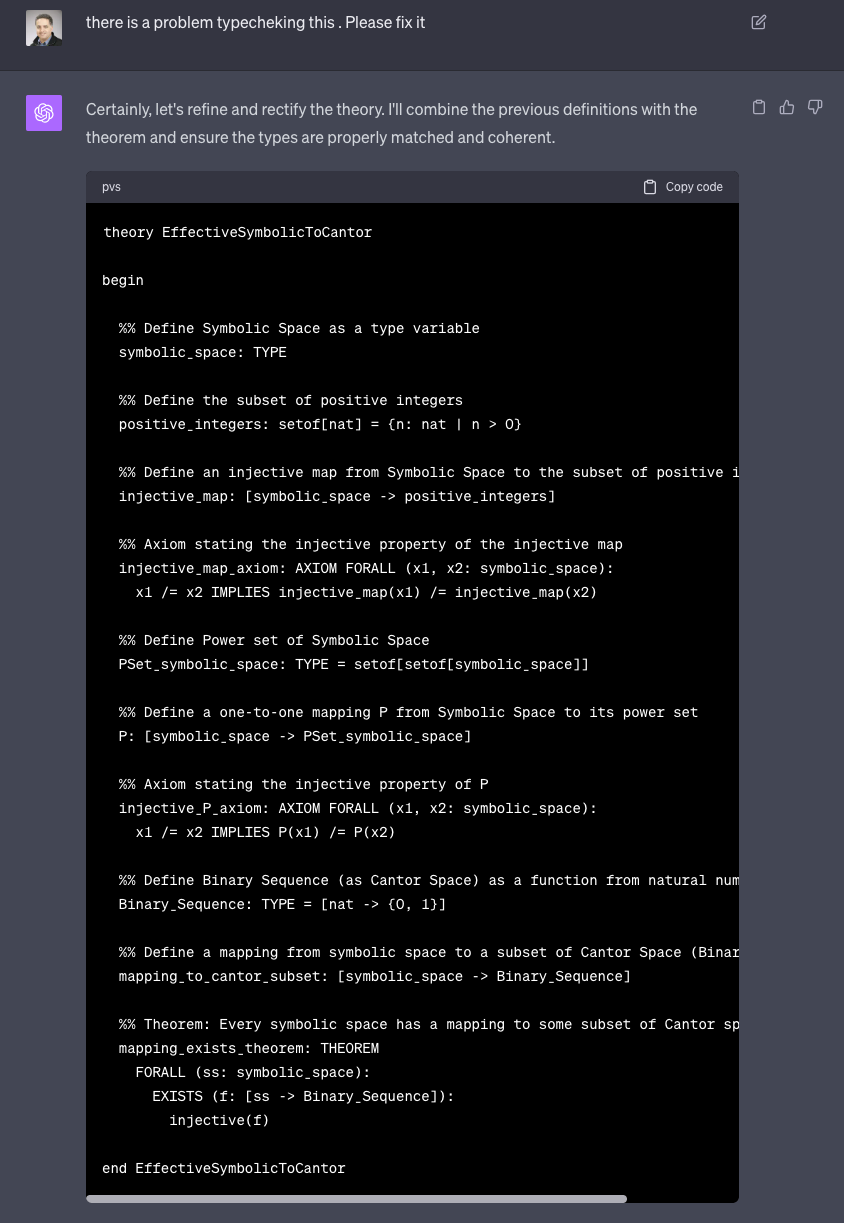}



\section*{Supplementary material}

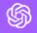
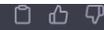

Let's break down the given definitions and theorem to understand them better:

**Definition 1: Effective Symbolic Space**

An effective symbolic space is defined as a symbolic space that has some specific properties. Symbolic spaces are mathematical structures that resemble topological spaces, but with symbols.

1. $X$ is a symbolic space.
2. $P : \mathbb{N} \to 2^X$ is an injective function. Here, $2^X$ denotes the power set of $X$ (the set of all subsets of $X$).
3. The range of $P$ is the set of all clopen sets in $X$. (Clopen sets are sets that are both open and closed in a topological space).
4. Intersecting and taking the complement of clopen sets are computable operations, which means there are functions that can determine the result of these operations algorithmically.

**Definition 2: Effective Continuous Map**

If you have two effective symbolic spaces, an effective continuous map is defined between them if:

1. $h : X \to Y$ is continuous.
2. For every clopen set in $Y$, its inverse image in $X$ is also a clopen set. Plus, there's a computable function that tells which clopen set in $X$ corresponds to a clopen set in $Y$ under the map $h$.
3. If $h$ is bijective (i.e., a one-to-one correspondence), then it is called an effective homeomorphism, and the two effective symbolic spaces are said to be effectively homeomorphic.

**Theorem 1: Effective Homeomorphism with Cantor Space**

The Cantor space is a classical object in topology and is a totally disconnected, perfect, compact metric space. The theorem states:

↻ Regenerate

1. Every effective symbolic space can be matched up (in a way that preserves the structure)

> simplify these concepts as mappings from symbolic spaces to the cantor spaces by going through natural numbers and theor binary representation

Alright, let's simplify the concepts using the relationship between natural numbers and their binary representations, and map symbolic spaces to the Cantor space.

### Symbolic Space and Cantor Space
Imagine a sequence of symbols, like `0` and `1`. A symbolic space can be thought of as all possible sequences of these symbols. The Cantor space is similar: it's all infinite sequences of `0`s and `1`s. So, each point in the Cantor space can be thought of as a binary number.

### Effective Symbolic Space
Now, let's simplify the effective symbolic space: it's a special kind of symbolic space where we can pick out specific sets (clopen sets) using natural numbers. Think of a function that, for every natural number $n$, assigns a set of sequences in our space. Moreover, we can compute the intersection and complement of these clopen sets algorithmically.

### Mapping Symbolic Spaces to Cantor Space
Since the Cantor space is made up of sequences of `0`s and `1`s, we can think of it as representing binary decimals. For instance, the sequence `0.001011...` can be thought of as a point in the Cantor space.

When we talk about an effective symbolic space being "effectively homeomorphic" to a subset of the Cantor space, we're saying there's a way to match up the sequences in our symbolic space to sequences of `0`s and `1`s in the Cantor space. We do this via a special function that uses natural numbers as a bridge.

### Effective Continuous Map Simplified
An effective continuous map between two symbolic spaces means we can move from one space to the other without tearing or gluing. For effective symbolic spaces, this is made concrete by using natural numbers and their binary representations. If you give me a set (represented by a number) in one space, I can tell you its counterpart set in the other space using a computable function.

### Connection to Binary Representation
The Cantor space's connection to binary numbers is direct since each point corresponds to a binary sequence. For effective symbolic spaces, the role of natural numbers (and by extension, their binary representations) is crucial. They provide the indexing mechanism to keep track of special sets (clopen sets) and mappings between spaces.

### The Big Picture Simplified
Imagine two huge books: one is the book of sequences from our symbolic space, and the other is the book of binary numbers. The theorem tells us that for every page in our first book, there's a matching page in the binary book. If our symbolic space is "perfect," then our first book is as thick as the binary book, meaning every binary number has a partner in our symbolic space. Otherwise, our symbolic space book might be thinner, but still, every in it finds a match in the binary book.

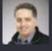 can you describe the relationships between all these concepts as a graph?

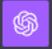 Certainly! Describing these relationships in a graph format will be textually based since I can't draw an actual graph here.

Imagine a flowchart or a network of nodes connected by arrows, each representing a concept and the connections between them:

1. **Symbolic Space (X)**
   - This is the starting node. Think of it as a box filled with sequences of symbols.
2. **Effective Symbolic Space (X, P)**
   - This is connected by an arrow from the Symbolic Space node.
   - Criteria for transition: There must exist a function $P$ that assigns clopen sets to natural numbers in a unique way.
3. **Cantor Space**
   - A standalone node, visualizing the space of all infinite binary sequences.
4. **Effective Subset of Cantor Space**
   - An arrow from Effective Symbolic Space points to this node.
   - This represents the idea that any effective symbolic space can be matched with some subset of the Cantor space.
5. **Perfect Effective Symbolic Space**
   - An arrow from Effective Symbolic Space also points to this, indicating that an effective symbolic space can have a special property of being "perfect."
6. **Effective Homeomorphism**
   - Arrows connect this node both to Effective Symbolic Space and Cantor Space.
   - It implies that when there exists an effective homeomorphism, there's a special one-to-one correspondence between the two spaces.
7. **Continuous Map (h)**
   - Connects two symbolic spaces (not necessarily different). It's a general concept that some spaces can be mapped to others in a continuous fashion.
8. **Effective Continuous Map**
   - An arrow from Continuous Map points to this, suggesting it's a specialized version.
   - Criteria for transition: The inverse image of any clopen set in the target space should be a clopen set in the original space, and there's a computable function to determine this relationship.
9. **Natural Numbers & Binary Representation**
   - A foundational node that connects to Effective Symbolic Space, Effective Continuous Map, and Cantor Space.
   - It serves as the bridge or the computational mechanism that powers the "effective" part of our constructs, making algorithmic operations possible.

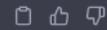

**Summary of Relationships:**

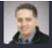
**make a textual graph**

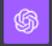
Alright, let's try to represent the relationships textually in a simple hierarchy using indentation:

```mathematica
Symbolic Space (X)
│
└── Effective Symbolic Space (X, P)
    │
    ├── Cantor Space
    │   │
    │   └── Effective Subset of Cantor Space
    │
    ├── Perfect Effective Symbolic Space
    │   │
    │   └── Cantor Space
    │
    ├── Continuous Map (h)
    │   │
    │   └── Effective Continuous Map
    │       │
    │       └── Natural Numbers & Binary Representation
    │
    └── Effective Homeomorphism
        │
        └── Natural Numbers & Binary Representation
```

In this textual representation:

- The main categories or concepts are on the left.
- Indentation implies dependency or a subset relationship.
- The connection lines (│ and └──) represent a relationship or progression from one concept to the next.

For simplicity, some relationships are represented in a linear hierarchy, though in a full graphical representation, there would be more interconnections between nodes. This textual format gives a basic overview of the relationships between the concepts.

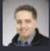 here is an abstraction of these concepts. Is this abstraction accurate? Definition 3. Given two
sets X and Y , a mapping h is a one-to-one mapping between its elements
(h(x) = y) such that x ∈ X and y ∈ Y and the inverse mapping exists h−1(y) = x.
Definition 4. A Cantor space is the set of infinite numbers on the interval [0, 1] that can be represented
as a sequence a0, a1, . . . such that ai ∈ {0, 1}
Definition 5. A symbolic space is a set such that there exists a one-to-one map to some subset of
integers greater than 0.
Definition 6. An effective symbolic space is a pair (X, P ) where X is a symbolic space and P is a
one-to-one mapping P : X → 2X . Here 2X is the power set of all possible combinations of subsets
of X.
Theorem. Every effective symbolic space has a homeomorphic mapping to some subset of a cantor
spac

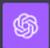 Your abstraction is an attempt to simplify and represent the previously discussed concepts in a clearer and more foundational way. Let's analyze each definition and the theorem to see if they capture the essence of the original concepts:

**Definition 3:** This definition describes a bijective mapping (one-to-one correspondence) between sets $X$ and $Y$. This is correct.

**Definition 4:** This captures the essence of the Cantor space, which is a set of all infinite binary sequences. By emphasizing sequences with entries in $\{0, 1\}$, it's a good representation.

**Definition 5:** This definition aims to define a symbolic space based on a mapping to positive integers. While symbolic spaces in the initial discussion were sequences of symbols, this abstraction does hint at the idea that each sequence can be uniquely identified, possibly hinting at an ordinal or indexing. However, the definition might be slightly restrictive. The essence of a symbolic space is more about the structure of sequences than their direct correspondence to integers.

**Definition 6:** This is an attempt to simplify the idea of an effective symbolic space. The definition hints at the computability essence of the original definition by introducing a mapping $P : X \to 2^X$, although the nature of $P$ being computable isn't directly mentioned.

**Theorem:** The theorem simplifies the original statement that every effective symbolic space is effectively homeomorphic to some subset of the Cantor space. It captures the primary essence, although the term "effectively homeomorphic" was more specific in its computable nature.

In summary, your abstraction does capture the general idea of the original concepts but misses some nuances. For more detailed discussions or applications, some of these nuances may be crucial. However, for a general overview or introductory explanation, this abstraction can be helpful.

↻ Regenerate

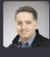
Given two sets $X$ and $Y$, a mapping $h$ is a one-to-one mapping between its elements ($h(x) = y$) such that $x\in X$ and $y\in Y$ and the inverse mapping exists $h^{-1}(y) = x$.

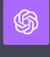
Yes, you are correct. A mapping $h : X \to Y$ is one-to-one (or injective) if it assigns to every element $x$ in set $X$ a unique element $y$ in set $Y$, such that if $h(x_1) = h(x_2)$, then $x_1 = x_2$. In other words, different elements in set $X$ are mapped to different elements in set $Y$.

Moreover, if there exists an inverse mapping $h^{-1} : Y \to X$, this means that the original mapping $h$ is bijective (both injective and surjective), meaning that it is one-to-one and onto. An "onto" (or surjective) mapping is one where every element in $Y$ is mapped to by some element in $X$. When a bijective mapping exists, it sets up a perfect one-to-one correspondence between the elements of sets $X$ and $Y$, allowing us to reverse the mapping and go from elements in $Y$ back to elements in $X$ through the inverse function $h^{-1}$. The existence of an inverse mapping inherently requires the original mapping to be bijective.

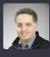
generate a PVS theory for this summary

From this point, we ask ChatGPT4 to generate PVS theories formalizing the various definitions and theorems. The resulting PVS theories are manually edited to produce a single theory merging the various generated theories. In particular, (1) fix syntax errors in the generated PVS code, and (2) change the theorem from proving a surjective mapping to a bijective mapping, and (3) added two lemmas that allow us to prove the main result.

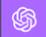 Alright, here's a single PVS theory that combines the definitions and properties of injective, surjective, bijective mappings, and their inverses.

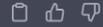

```pvs
theory Mappings
IMPORTING set_theory %% Assuming a set_theory with basic set definitions and

begin

var
  X: TYPE,
  Y: TYPE

%% Define Mapping
h: [X -> Y]

%% Define Injective (One-to-One) Mapping
injective(h): bool = FORALL (x1, x2: X): h(x1) = h(x2) IMPLIES x1 = x2

%% Define Surjective (Onto) Mapping
surjective(h): bool = FORALL (y: Y): EXISTS (x: X): h(x) = y

%% Define Bijective Mapping
bijective(h): bool = injective(h) AND surjective(h)

%% Define Inverse Mapping
h_inv: [Y -> X]

%% Define property of Inverse Mapping
inverse_mapping(h, h_inv): bool = FORALL (x: X, y: Y): (h(x) = y IMPLIES h_i

%% Lemma stating that if a mapping has an inverse, then it is bijective
bijective_if_has_inverse: LEMMA
  FORALL (h, h_inv: [X -> Y]):
    inverse_mapping(h, h_inv) IMPLIES bijective(h)

end Mappings
```

This PVS theory provides a simplified abstract view of injective, surjective, and bijecti...

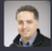
A Cantor space is the set of infinite numbers that can be represented as a sequence $a_0, a_1, \hdots$ such that $a_i \in \{0,1\}$

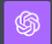
A Cantor space is indeed related to the set of all infinite binary sequences, where each element $a_i$ of a sequence is in the set $\{0, 1\}$. Formally, the Cantor space is the product of countably many copies of the discrete two-point space $\{0, 1\}$, and it can be represented as $\{0, 1\}^\omega$, where $\omega$ denotes a countable infinity, corresponding to the sequence indices.

## PVS Theory

Here's how you might begin to define a Cantor space in PVS.

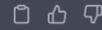
```pvs
theory CantorSpace

begin

  %% Define the Binary Sequence Type as a function from natural numbers to {
  Binary_Sequence: TYPE = [nat -> {0, 1}]

  %% Cantor space is the set of all binary sequences.
  Cantor_Space: TYPE = setof[Binary_Sequence]

  %% Example of an element in Cantor space: the sequence 01010101...
  alternating_sequence: Binary_Sequence = (LAMBDA (n: nat): IF even(n) THEN

end CantorSpace
```

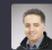
\begin{theorem}
 Every effective symbolic space has a mapping to some subset of a cantor space.
\end{theorem}

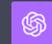
The statement you've presented is a theorem asserting that for every effective symbolic space, there exists a mapping to some subset of a Cantor space. To understand this, we should note that a Cantor space is essentially the space of all infinite binary sequences, and it has the cardinality of the continuum.

An effective symbolic space, by definition, includes a symbolic space $X$, which has a one-to-one mapping to a subset of positive integers, thus it is countable. The Cantor space is uncountable; therefore, every countable set (including every symbolic space) can indeed be mapped injectively into the Cantor space. This mapping can be constructed in various ways, depending on the specificities of the effective symbolic space and the desired properties of the mapping.

Here's a simplified illustrative representation in PVS of this theorem, building on the previous definitions:

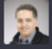 there is a problem typecheking this . Please fix it

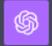 Certainly, let's refine and rectify the theory. I'll combine the previous definitions with the theorem and ensure the types are properly matched and coherent.

```pvs
theory EffectiveSymbolicToCantor

begin

  %% Define Symbolic Space as a type variable
  symbolic_space: TYPE

  %% Define the subset of positive integers
  positive_integers: setof[nat] = {n: nat | n > 0}

  %% Define an injective map from Symbolic Space to the subset of positive i
  injective_map: [symbolic_space -> positive_integers]

  %% Axiom stating the injective property of the injective map
  injective_map_axiom: AXIOM FORALL (x1, x2: symbolic_space):
    x1 /= x2 IMPLIES injective_map(x1) /= injective_map(x2)

  %% Define Power set of Symbolic Space
  PSet_symbolic_space: TYPE = setof[setof[symbolic_space]]

  %% Define a one-to-one mapping P from Symbolic Space to its power set
  P: [symbolic_space -> PSet_symbolic_space]

  %% Axiom stating the injective property of P
  injective_P_axiom: AXIOM FORALL (x1, x2: symbolic_space):
    x1 /= x2 IMPLIES P(x1) /= P(x2)

  %% Define Binary Sequence (as Cantor Space) as a function from natural num
  Binary_Sequence: TYPE = [nat -> {0, 1}]

  %% Define a mapping from symbolic space to a subset of Cantor Space (Binar
  mapping_to_cantor_subset: [symbolic_space -> Binary_Sequence]

  %% Theorem: Every symbolic space has a mapping to some subset of Cantor sp
  mapping_exists_theorem: THEOREM
    FORALL (ss: symbolic_space):
      EXISTS (f: [ss -> Binary_Sequence]):
        injective(f)

end EffectiveSymbolicToCantor
```

Mappings: THEORY
  BEGIN

  IMPORTING sets

  $X$: TYPE

  $Y$: TYPE

  $H$: TYPE = $[X \rightarrow Y]$

  injective($h$: $H$): bool =
    $\forall$ ($x_1$, $x_2$: $X$): $h(x_1) = h(x_2)$ IMPLIES $x_1 = x_2$

  surjective($h$: $H$): bool =
    $\forall$ ($y$: $Y$): $\exists$ ($x$: $X$): $h(x) = y$

  bijective($h$: $H$): bool = injective($h$) AND surjective($h$)

  H_INV: TYPE = $[Y \rightarrow X]$

  inverse_mapping($h$: $H$, h_inv: H_INV): bool =
    $\forall$ ($x$: $X$, $y$: $Y$):
      ($h(x) = y$ IMPLIES h_inv($y$) = $x$) AND
      (h_inv($y$) = $x$ IMPLIES $h(x) = y$)

  bijective_if_has_inverse: LEMMA
    $\forall$ ($h$: $H$, h_inv: H_INV):
      injective($h$) AND inverse_mapping($h$, h_inv) IMPLIES bijective($h$)

  ZeroOne: TYPE = $\{x: \mathbb{N} \mid x = 0$ OR $x = 1\}$

  Binary_Sequence: TYPE = $[\mathbb{N} \rightarrow$ sequence[ZeroOne]$]$

  Cantor_Space: TYPE = setof[Binary_Sequence]

  CS: Cantor_Space

  symbolic_space_element: TYPE

  symbolic_space: TYPE = setof[symbolic_space_element]

$\mathbb{N}_{>0}$: TYPE = $\{x: \mathbb{N} \mid x > 0\}$

positive_integers: TYPE = setof$[\mathbb{N}_{>0}]$

injective_map: [symbolic_space $\to$ positive_integers]

injective_map_axiom: AXIOM
  $\forall$ ($x_1$, $x_2$: symbolic_space):
    $x_1 \neq x_2$ IMPLIES injective_map($x_1$) $\neq$ injective_map($x_2$)

PSet_symbolic_space: TYPE = setof[setof[symbolic_space]]

$P$: [symbolic_space $\to$ PSet_symbolic_space]

injective_P_axiom: AXIOM
  $\forall$ ($x_1$, $x_2$: symbolic_space): $x_1 \neq x_2$ IMPLIES $\mathbb{P}(x_1) \neq \mathbb{P}(x_2)$

effective_symbolic_space: TYPE =
    [#ss_field: symbolic_space, pss_field: PSet_symbolic_space#]

effective_symbolic_space_axiom(ss: symbolic_space): effective_symbolic_space =
  (#ss_field := ss, pss_field := $\mathbb{P}$(ss)#)

subset_Cantor_Space: TYPE = setof[Cantor_Space]

injective($f$: [effective_symbolic_space $\to$ subset_Cantor_Space]): bool =
  $\forall$ ($x_1$, $x_2$: effective_symbolic_space):
    $x_1 \neq x_2$ IMPLIES $f(x_1) \neq f(x_2)$

bijective_map($f$: [symbolic_space $\to$ subset_Cantor_Space]): bool =
  $\forall$ ($x_1$, $x_2$: symbolic_space): $x_1 = x_2$ IMPLIES $f(x_1) = f(x_2)$

MapLemma1: LEMMA $\exists$ ($f$: [symbolic_space $\to \mathbb{N}$]): bijective?($f$)

MapLemma2: LEMMA $\exists$ ($f$: [$\mathbb{N} \to$ Binary_Sequence]): bijective?($f$)

mapping_exists_theorem: THEOREM
  $\forall$ (ss: symbolic_space):
    $\exists$ ($f$: [symbolic_space $\to$ Binary_Sequence]): bijective?($f$)

END Mappings

Verbose proof for `mapping_exists_theorem`.
`mapping_exists_theorem`:

| {1} | ∀ (ss: symbolic_space): ∃ (f: [symbolic_space → Binary_Sequence]): bijective?(f) |

`mapping_exists_theorem`:

| [1] | ∀ (ss: symbolic_space): ∃ (f: [symbolic_space → Binary_Sequence]): bijective?(f) |

Rule? (lemma "MapLemma1")
Applying MapLemma1
`mapping_exists_theorem`:

| {-1} | ∃ (f: [symbolic_space → ℕ]): bijective?(f) |
| [1] | ∀ (ss: symbolic_space): ∃ (f: [symbolic_space → Binary_Sequence]): bijective?(f) |

Rule? (lemma "MapLemma2")
Applying MapLemma2
`mapping_exists_theorem`:

| {-1} | ∃ (f: [ℕ → Binary_Sequence]): bijective?(f) |
| [-2] | ∃ (f: [symbolic_space → ℕ]): bijective?(f) |
| [1] | ∀ (ss: symbolic_space): ∃ (f: [symbolic_space → Binary_Sequence]): bijective?(f) |

Rule? (skeep)
Skolemizing and keeping names of the universal formula in (+ -),
`mapping_exists_theorem`:

| [-1] | ∃ (f: [ℕ → Binary_Sequence]): bijective?(f) |
| [-2] | ∃ (f: [symbolic_space → ℕ]): bijective?(f) |
| {1} | ∃ (f: [symbolic_space → Binary_Sequence]): bijective?(f) |

Rule? (skeep)
Skolemizing and keeping names of the universal formula in (+ -),
`mapping_exists_theorem`:

| {-1} | bijective?(f) |
| [-2] | ∃ (f: [symbolic_space → ℕ]): bijective?(f) |
| [1] | ∃ (f: [symbolic_space → Binary_Sequence]): bijective?(f) |

Rule? (skeep)
Skolemizing and keeping names of the universal formula in (+ -),

1

```
mapping_exists_theorem :

  [-1]  bijective?(f)
  {-2}  bijective?(f′)
  ──────────────────────────────────────────────
  [1]   ∃ (f: [symbolic_space → Binary_Sequence]): bijective?(f)
```

Rule? (inst 1 "f o f!1")
Instantiating the top quantifier in 1 with the terms: $f \circ f'$,
```
mapping_exists_theorem :

  [-1]  bijective?(f)
  [-2]  bijective?(f′)
  ──────────────────────────────────────────────
  {1}   bijective?(f ∘ f′)
```

Rule? (assert)
Simplifying, rewriting, and recording with decision procedures,
```
mapping_exists_theorem :

  [-1]  bijective?(f)
  [-2]  bijective?(f′)
  ──────────────────────────────────────────────
  [1]   bijective?(f ∘ f′)
```

Rule? (use "composition_bijective[symbolic_space,nat,Binary_Sequence]")
Using lemma composition_bijective[symbolic_space, ℕ, Binary_Sequence],
This completes the proof of `mapping_exists_theorem`.
Q.E.D.



\end{document}